\documentclass[smallcondensed]{svjour3}     
\smartqed  

\usepackage{booktabs}
\usepackage{graphicx}
\usepackage{amsfonts} 
\usepackage{amssymb} 
\usepackage{xspace} 
\usepackage{xfrac}
\usepackage{xcolor}
\usepackage{url}
\usepackage{makeidx}
\usepackage{multirow}
\usepackage{amsmath}
\usepackage{footnote}
\usepackage[misc]{ifsym}

\begin{document}

\title{Multi-output Ensembles for Multi-step Forecasting
}

\titlerunning{Multi-output Ensembles for Multi-step Forecasting}        

\author{Vitor~Cerqueira         \and
        Luis~Torgo
}

\authorrunning{V. Cerqueira et al.} 

\institute{V. Cerqueira (\Letter) \at
         Dalhousie University, Halifax, Canada\\
         \email{vitor.cerqueira@dal.ca}
         \and
         Lu\'{i}s Torgo \at
         Dalhousie University, Halifax, Canada\\
         \email{ltorgo@dal.ca}
}

\date{Received: date / Accepted: date}

\maketitle

\begin{abstract}

This paper studies the application of ensembles composed of multi-output models for multi-step ahead forecasting problems.
Dynamic ensembles have been commonly used for forecasting. However, these are typically designed for one-step ahead tasks.
On the other hand, the literature regarding the application of dynamic ensembles for multi-step ahead forecasting is scarce. Moreover, it is not clear how the combination rule is applied across the forecasting horizon.
We carried out extensive experiments to analyze the application of dynamic ensembles for multi-step forecasting. We resorted to a case study with 3568 time series and an ensemble of 30 multi-output models.
We discovered that dynamic ensembles based on arbitrating and windowing present the best performance according to average rank. Moreover, as the horizon increases, most approaches struggle to outperform a static ensemble that assigns equal weights to all models.
The experiments are publicly available in a repository.

\end{abstract}

\section{Introduction}\label{intro}

Ensemble methods combine the output of several models to make aggregated predictions. These approaches have been shown to improve predictive performance over single models in many tasks \cite{brown2005managing}, including forecasting \cite{cerqueira2017dets}. One of the reasons to use an ensemble is that it reduces the chance of selecting the wrong model \cite{hibon2005combine}.

Ensembles have attracted much attention in the literature. Bates and Granger \cite{bates1969combination} pioneered the combination of different models for time series forecasting. Since then, hundreds of works have been devoted to this topic. 
The simplest approach to forecast combination is to weigh each model equally by averaging the prediction with the arithmetic mean. Weighted averages can also be used, where the weights reflect how well we expect each model to perform.

Time series are often non-stationary. This means that the data distribution changes over time. Therefore, the combination rule for forecasting problems is typically dynamic. The weights change over time to adapt to changes in the time series.

There are many different forecast combination strategies in the literature. Most approaches are based on regret minimization \cite{Cesa-Bianchi:2006:PLG:1137817}, windowing \cite{Blast}, or meta-learning \cite{cerqueira2019arbitrage}.
However, dynamic approaches are typically designed for one-step ahead forecasting problems. They assume that there is immediate feedback concerning the actual value for computing the error and updating the combination rule.

Forecasting multiple observations in advance is important for many application domains. It helps to reduce the long-term uncertainty of time series. Consequently, this results in better planning within organizations. Approaches for multi-step forecasting include the recursive, direct, or multi-output methods \cite{taieb2012review}. Multi-output methods have been shown to provide better forecasting performance in multi-step ahead prediction settings \cite{taieb2012review}. In effect, our work focuses on this type of approach. 

However, since dynamic combination methods are usually designed for one-step ahead settings, it is not clear how they should be applied for multi-step forecasting. 
For example, one could compute different weights for each horizon. Or estimate the weights for all forecasting horizons jointly.

In this work, we investigate the performance of ensembles composed of multi-output models for multi-step forecasting problems. 
Our goal is two-fold. First, we aim at studying which dynamic combination rule is the best for multi-step ahead problems. Previous works have carried out such analysis for one-step ahead tasks, e.g. \cite{cerqueira2019arbitrage}. Conversely, the literature on dynamic ensembles for multi-step ahead forecasting is scarce.
The second objective is to analyze which is the best approach for computing the ensemble weights along the forecasting horizon. To our knowledge, no prior work has addressed this topic.

The analysis presented in this work is based on an empirical study. The experiments included a set of 3568 univariate time series from multiple domains. We test different combination rules using an ensemble composed of 30 individual multi-output models. These models are standard supervised learning regression algorithms.

The results suggest that the dynamic combination rules based on arbitrating \cite{cerqueira2019arbitrage} and windowing \cite{newbold1974experience} present the best average rank. Yet, most approaches struggle to outperform the combination rule which assigns equal weights to all models. We discovered that the forecasting horizon has a significant impact on these results. Dynamic combination rules tend to perform better for short-term forecasting. On the other hand, these approaches decrease their performance as the forecasting horizon increases.
In terms of computing the ensemble weights along the horizon, no considerable differences were found between the tested approaches. 
The data sets and experiments are publicly available in an online repository\footnote{\url{https://github.com/vcerqueira/mo_ensembles}}.

The rest of the paper is organized as follows. In the next section (Section \ref{sec:rw}), we provide a background to our work. We start by defining the time series predictive task. We also overview the literature related to our work. We focus on the topics of dynamic ensembles and multi-step forecasting.
In Section \ref{sec:materialsmethods}, we present the materials and methods used in this paper. We describe the case study which contains time series from several application domains. We also describe the methods used in detail. Moreover, we explain the experimental design used in the experiments.
Then, we present the experiments in Section \ref{sec:ee}. In that section, we outline the research questions and provide an empirical answer to them.
We discuss the results obtained in Section \ref{sec:disc}, pointing out some future directions. Finally, the paper is concluded in Section \ref{sec:fr}.

\section{Background}\label{sec:rw}

We split the literature related to our work into five parts: time series forecasting (Section \ref{sec:tsf}), ensemble methods (Section \ref{sec:rw1}), dynamic ensembles for forecasting (Section \ref{sec:rw2}), ensemble pruning (Section \ref{sec:pruning}), and multi-step ahead forecasting (Section \ref{sec:rw3}).

We start by formalizing the problem of time series forecasting in Section \ref{sec:tsf}.
Section \ref{sec:rw1} outlines the motivation for using ensemble methods. There, we refer to seminal works on this topic.
Section \ref{sec:rw2} overviews the application of ensemble methods to forecasting problems. We focus on a particular type of ensemble, namely dynamic and heterogeneous ensembles. These are commonly used to tackle forecasting tasks. Within Section \ref{sec:rw2}, we also list several methods that are split into three categories: windowing (Section \ref{sec:rw2.1}), regret minimization (Section \ref{sec:rw2.2}), and meta-learning (Section \ref{sec:rw2.3}). Then, we describe ensemble pruning in Section \ref{sec:pruning}, which is an important step for improving forecasting performance.
Finally, we list state-of-the-art approaches to multi-step ahead forecasting in Section \ref{sec:rw3}. These are split into two types: single-output methods and multi-output methods. We explain the difference between the two and our choice for the second type.

\subsection{Time Series Forecasting}\label{sec:tsf}

We define a univariate time series as a sequence of values $Y = \{y_1, y_2, \dots,$ $y_n \}$, where $y_i \in \mathcal{Y} \subset \mathbb{R}$ is the numeric value of the time series at time $i$ and $n$ is the length of $Y$. We assume that the observations of the series are captured at regular periods (e.g. every hour).
This work addresses multi-step ahead forecasting (also referred to as long-term forecasting \cite{taieb2012review}) problems in univariate time series.
In practice, we aim at predicting the value of the $H$ upcoming observations of the time series, $y_{n+1}, \ldots, y_{n+H}$, where $H$ denotes the forecasting horizon. 

The predictive task is formalized as an auto-regressive problem. Thus, each observation is represented based on the past and recent values before it. This is accomplished by reconstructing the time series using time delay embedding \cite{Takens1981}. This method is used to transform the time series from a sequence to a tabular format.

The transformation based on time delay embedding leads to a data set $\mathcal{D}=\{$($X$, $y$)$\}$. 
As described before, each observation $y_i$ is modeled according to past $q$ values before it: $X_i = \{y_{i-1}, y_{i-2}, \dots, y_{i-q} \}$, where $y_i \in \mathcal{Y} \subset \mathbb{R}$. In effect, $y$ represents the target variable
which represents the observation we want to predict, and $X_i \in \mathcal{X} \subset \mathbb{R}^q$ represents the $i$-th embedding vector. 
Then, we train a multiple regression model $f$ that can be written as $y_i = f(X_i)$.
This definition is designed for one-step ahead forecasting. In Section \ref{sec:rw3}, we describe how this formalization is extended to the multi-step ahead case.

\subsection{Ensemble Methods}\label{sec:rw1}

The \textit{No Free Lunch} theorem for supervised learning \cite{wolpert1996lack} postulates that no learning algorithm is the most appropriate for all problems. All methods have strengths and limitations. This is the key motivation for ensemble methods \cite{brown2005managing}.

Ensemble methods aim at combining the output of several different predictive models. These methods have been shown to perform better than individual models in many tasks and domains of applications.

A key aspect of developing accurate ensembles is the diversity among individual models. The models composing the ensemble should be accurate but different from each other. Brown et al. \cite{brown2005diversity} survey several methods to achieve this. 
In this paper, we focus on heterogeneous ensembles \cite{kuncheva2004classifier}. These ensembles are composed of individual models which are trained with distinct learning algorithms, which is a typical way of encouraging diversity.

\subsection{Dynamic Ensembles for Forecasting}\label{sec:rw2}

Time series forecasting is one of the tasks in which ensembles have been shown to provide state-of-the-art performance \cite{bates1969combination,cerqueira2019arbitrage,oreshkin2019n}. Hibon and Evgeniou \cite{hibon2005combine} argue that combining multiple different forecasting models reduces the chance of picking the wrong individual model. 

Time series data is amenable to change due to sources of non-stationarity. Consequently, different forecasting models usually show varying relative performance \cite{aiolfi2006persistence}. Aiolfi and Timmermann study this phenomenon. They discovered that some models perform better than others in some periods over a time series. This is the main motivation for using dynamic ensembles for forecasting.

A dynamic ensemble combines the predictions of individual models using weighted averages, where the weights change over time. The weights change to adapt to the current process generating the time series. The main problem when using dynamic ensembles is defining how to compute these weights at each time-step. We overview several approaches to accomplish this. These are split into three types:
windowing approaches (Section \ref{sec:rw2.1}), regret minimization approaches (Section \ref{sec:rw2.2}), and meta-learning approaches (Section \ref{sec:rw2.3}).

\subsubsection{Windowing Approaches}\label{sec:rw2.1}

The static forecast combination approach which assigns equal weights to all available models is a robust combination method \cite{clemen1986combining} (\texttt{Simple}). Another static combination approach is a weighted average, in which the weights are set
according to the performance of models in the training data (\texttt{LossTrain}). A variant of this approach is to select the model with the best performance on the training data (\texttt{Best}). Notwithstanding, dynamic combination rules are also typically used in forecasting problems.

Using past recent performance is a typical way of dynamically combining ensembles. The idea is to give a higher weight to models that performed better in recent observations (\texttt{Window}).
This approach has shown promising results in different works \cite{newbold1974experience,cerqueira2017dynamic}.
A variant of this is the method called \texttt{Blast}, which selects the best model in recent observations for forecasting \cite{Blast}.

An alternative to computing performance in a window of recent data is to use forgetting factors. These take into account all past observations but give more relevance to more recent ones. An example method that follows this approach is \texttt{AEC} \cite{sanchez2008adaptive}.

\subsubsection{Regret Minimization}\label{sec:rw2.2}

Combining the output of multiple models is a well-studied topic in the online learning literature \cite{Cesa-Bianchi:2006:PLG:1137817}. Example methods include the exponentially weighted average (\texttt{EWA}), the polynomially weighted average (\texttt{MLpol}), and the fixed share aggregation (\texttt{FS}). 

These approaches are designed to minimize regret. Regret is the average error suffered relative to the best error we could have obtained. We refer the reader to the second chapter of the seminal work by Cesa-Bianchi and Lugosi \cite{Cesa-Bianchi:2006:PLG:1137817}.

\subsubsection{Meta-learning}\label{sec:rw2.3}

Meta-learning is another type of strategy that can be used to combine the output of multiple models.
Arguably, the most popular meta-learning approach is \texttt{Stacking}, which was proposed by Wolpert \cite{wolpert1992stacked}.

Cerqueira et al. \cite{cerqueira2019arbitrage} proposed a meta-learning approach called Arbitrated Dynamic Ensemble (\texttt{ADE}) for dynamic forecast combinations.
\texttt{ADE} works by building a meta-model for each model in the ensemble. Each meta-model is designed to model and forecast the error of the corresponding (base) model. Then, the models in the ensemble are weighted according to the error forecasts provided by meta-models. 

\subsection{Ensemble Pruning}\label{sec:pruning}

Pruning is an important stage in the creation of ensembles \cite{mendes2012ensemble}. The goal of this process is to remove poor-performing models from the ensemble.

Pruning has been shown to improve the performance of forecasting ensembles \cite{jose2008simple,cerqueira2019arbitrage}. Jose and Winkler \cite{jose2008simple} trim a percentage of the worst models based on past performance. We also take this approach in this work.

\subsection{Multi-step Forecasting}\label{sec:rw3}

Multi-step ahead forecasting denotes the process of predicting multiple instances of a time series. This task reduces the long-term uncertainty of time series, which is desirable across many application domains. Several works have been devoted to multi-step ahead forecasting problems \cite{taieb2012review,venkatraman2015improving,de2019batch}.

In this section, we review several state-of-the-art approaches for multi-step ahead forecasting. We split these into two types: single-output methods (Section \ref{sec:rw_so}), and multi-output methods (Section \ref{sec:rw_mo}). Taieb et al. \cite{taieb2012review} are followed closely to describe some of these methods. 

\subsubsection{Single Output Methods}\label{sec:rw_so}

One of the most popular approaches to multi-step ahead forecasting is the Recursive method (also known as the Iterative). 
Recursive works by fitting a model $f$ for one-step ahead forecasting (c.f. Section \ref{sec:tsf}): $$y_{i+1} = f(\{y_{i}, y_{i-1}, \dots, y_{i-q+1} \})$$ 
To get predictions for the next $H$ observations, the model $f$ is iterated $H$. To be more precise, in the $i$-th time-step, $\{y_{i}, y_{i-1}, \dots, y_{i-q+1} \}$ is the input used to forecast the value of $y_{i+1}$. Let $\hat{y}_{i+1}$ denote that prediction. Then, $\{\hat{y}_{i+1}, y_{i}, \dots, y_{i-q} \}$ is the input vector for forecasting $\hat{y}_{i+2}$ in the same time-step.

The Recursive strategy is known to propagate errors along the forecasting horizon. Thus, this approach requires an accurate model specification to work well.

Some methods have been proposed to mitigate the error propagation problem of Recursive. 
The method \texttt{RECNOISY}, proposed by Taieb and Bontempi \cite{taieb2011recursive}, adds noise to the input data at each forecasting horizon. The idea behind this approach is to make the forecasting model robust to error propagation. 

The method Data as Demonstrator (\texttt{DaD}) by Venkatraman et al. \cite{venkatraman2015improving,venkatraman2016improved} is also specifically designed to deal with the error propagation problem.
The idea is to use the training set to correct errors that occur during multi-step prediction. This is accomplished by iteratively augmenting the training set with these corrections. After the augmentation process terminates, the Recursive approach is applied. The augmented training data is used to train the forecasting model.

Another popular multi-step ahead forecasting method is the Direct approach. This strategy works by building a forecasting model for each horizon: $$y_{i+h} = f_h(\{y_{i}, y_{i-1}, \dots, y_{i-q+1} \}),$$ where $h \in \{1, \dots, H\}$.
Since no model is iterated along the horizon, the Direct approach does not suffer from error propagation.
Notwithstanding, this method leads to greater computational costs because it trains a model for each horizon. Besides, it assumes that each horizon is independent, which is not generally true.

The method DirRec attempts to bridge the best aspects of Recursive and Direct.
Similarly to Direct, this approach builds one model for each forecasting horizon. Moreover, for each successive horizon, the set of inputs is augmented with the predictions from previous steps (following Recursive). This approach is also referred to as classifier (or regression) chains in the machine learning literature \cite{read2021classifier}.

\subsubsection{Multiple Output Methods}\label{sec:rw_mo}

As the name implied, single output approaches model one horizon at a time. Thus, they do not account for the stochastic dependency along the forecasting horizon.
This aspect is taken into account by multi-input multi-output (MIMO) models \cite{taieb2012review}. In this work, we also refer to these approaches by multi-output models. 

A multi-output model is trained on the complete forecasting horizon jointly: $$[y_{i+1}, y_{i+2}, \dots, y_{i+H}] = F(\{y_{i}, y_{i-1}, \dots, y_{i-q+1} \}),$$ where $F$ denotes the multi-output model.
Essentially, there are $H$ target variables instead of one, and these are modeled jointly by a single model.

Taieb et al. \cite{taieb2012review} compared several approaches for multi-step ahead forecasting, including the ones described above. They used the 111 time series from the NN5 forecasting competition. 
The main conclusion is that multi-output strategies perform better than single-output ones.
For this reason, we focus on multi-output methods in this work. Notwithstanding, the dynamic ensembles used in this paper can also be applied with other multi-step ahead forecasting strategies.

\section{Materials and Methods}\label{sec:materialsmethods}

This section details the materials and methods used in this work. 
We describe the data sets used in the experiments in Section \ref{sec:data}.
Then, we detail the methodology carried out to evaluate each approach (Section \ref{sec:methodology}). We list all approaches in Section \ref{sec:actmethods}, including learning algorithms, dynamic combination rules, and different strategies for weighting models along the horizon. Finally, we detail the experimental design in Section \ref{ssec:ed}, including the cross-validation procedure and evaluation metric.

\subsection{Data}\label{sec:data}

The experiments encompassed 3568 time series. These were collected from the following 7 popular databases:
\begin{itemize}
    \item \texttt{electricity\_nips}: 370 hourly time series concerning the electricity consumption of 370 customers \cite{yu2015high}; 
    \item \texttt{nn5\_daily\_without\_missing}: 111 daily time series about cash withdrawal amounts \cite{taieb2012review};
    \item \texttt{solar-energy} 137 hourly time series representing power produced from solar energy \cite{lai2018modeling}:
    \item \texttt{traffic\_nips}: 963 hourly time series concerning the occupancy rates (between zero and one) of car lanes in San Francisco bay area freeways \cite{yu2015high};
    \item \texttt{taxi\_30min}: 1214 time series related to the number of taxi rides in different locations around New York in 30-minute windows \cite{borchert2022multi};
    \item \texttt{m4\_hourly}: 414 hourly time series from the M4 time series database \cite{makridakis2020m4} which cover different application domains;
    \item \texttt{m4\_weekly}: 414 weekly time series from the M4 time series database \cite{makridakis2020m4} which cover different application domains.
\end{itemize}

Table \ref{tab:datasets} presents a summary of the data sets, including the number of time series and their average length. The name of the data sets refers to their identifier in the Python library \textit{gluonts}.

\begin{table}[!bth]
	\centering
	\caption{Summary of the data sets}		
	\begin{tabular}{llrr}
	\toprule
	\textbf{Name} & \textbf{Frequency} & \textbf{\# Time Series} & \textbf{Avg. Length}\\
	    \midrule
        
        \texttt{electricity\_nips} & Hourly & 370 & 5833 \\
        
        \midrule  
        
        \texttt{nn5\_daily\_without\_missing} & Daily & 111 & 735 \\
        
        \midrule  
        
        \texttt{solar-energy} & Hourly & 137 & 7009 \\
        
        \midrule  
        
        \texttt{traffic\_nips} & Hourly & 963 & 4001 \\
        
        \midrule

        \texttt{taxi\_30min} & Half-hourly & 1214 & 1488 \\
        
        \midrule
        
        \texttt{m4\_hourly} & Hourly & 414 & 960 \\
        
        \midrule
        
        \texttt{m4\_weekly} & Weekly & 359 & 934 \\
        
		\bottomrule    
	\end{tabular}%
	\label{tab:datasets}
\end{table}

\subsection{Methodology}\label{sec:methodology}

The goal of this paper is to analyze how several dynamic ensembles perform for multi-step ahead forecasting problems. This is accomplished by carrying out a set of experiments.

For each time series in the case study (c.f. Section \ref{sec:data}), we apply the following procedure. The available time series is split into training and test sets. As we describe below in Section \ref{ssec:ed}, this process is repeated using a cross-validation procedure.

The models in the ensemble are fit using the training data. As mentioned in Section \ref{sec:rw2.2}, pruning the ensemble usually leads to better forecasting performance. In effect, we prune the ensemble and keep only the best 75\% of models in the available pool. The pruning process is carried out using a nested validation procedure. The training set is further split into two parts: an inner training set and a validation set. The inner training set contains 70\% of the observations of the complete (outer) training set. The validation set contains the final 30\% of the complete training set. All models are fit using the inner training set and evaluated in the validation set. The 25\% of models with the worst forecasting performance are discarded. The remaining ones are re-fit using the complete training set.

After the pruning and re-fitting process, the ensemble is applied to the test set. A combination rule is applied to aggregate the predictions of all available models. The list of all combination rules applied in the experiments is described in Section \ref{sec:combs}. Note that we test these combination rules with different approaches concerning the estimation of weights across the horizon. We test four strategies that are listed in Section \ref{sec:weightfuncs}. Finally, the performance of each method is evaluated according to its performance in the test set. The evaluation metric is described in Section \ref{ssec:ed}.

\subsection{Methods}\label{sec:actmethods}

This section details the methods used in the experiments. First, we describe the regression learning algorithms used to create multi-output forecasting models (Section \ref{sec:algos}). Then, we describe the 9 methods used to combine the predictions of those models (Section \ref{sec:combs}). Finally, we detail four different approaches to compute the ensemble weights across the forecasting horizon (Section \ref{sec:weightfuncs}).

\subsubsection{Learning Algorithms}\label{sec:algos}

The ensembles used in the experiments are composed of 40 individual multi-output models. These were created using the following learning algorithms:  random forest regression, extra trees regression, bagging of decision trees, projection pursuit regression, LASSO regression, ridge regression, elastic-net regression, k-nearest neighbors regression, principal components regression, and partial least squares regression. We used the implementation in the \textit{scikit-learn} \cite{pedregosa2011scikit} Python library to use these methods. Table \ref{tab:expertsspecs} describes the different parameters used for each learning algorithm. In total, there are 40 different learning approaches.

k-nearest neighbors regression, principal components regression, and partial least squares regression.

\begin{table}[!bth]
	\centering
	\caption{Summary of the parameters of the learning algorithms}		
	\resizebox{.9\textwidth}{!}{%
	\begin{tabular}{llll}
	\toprule
	\textbf{ID} & \textbf{Algorithm} & \textbf{Parameter(s)} & \textbf{Value(s)}\\
	\midrule
	    \texttt{BAGGING\_1} & \multirow{2}{*}{Bagging of decision trees} & \multirow{2}{*}{No. trees} & 50\\
	    
	    \texttt{BAGGING\_2} &  &  & 100\\
	    
	    \midrule   
	    
	    \texttt{RF\_1} & \multirow{6}{*}{Random Forest} & \multirow{6}{*}{\{No. trees, max depth\}} & \{50, \textit{default}\} \\
	    
	    \texttt{RF\_2} & & & \{50, 3\} \\
	    
	    \texttt{RF\_3} & & & \{50, 5\} \\
	    
	    \texttt{RF\_4} & & & \{100, \textit{default}\} \\
	    
	    \texttt{RF\_5} & & & \{100, 3\} \\
	    
	    \texttt{RF\_6} & & & \{100, 5\} \\
	    
	    \midrule   
	    
	    \texttt{ET\_1} & \multirow{6}{*}{Extra trees regression} & \multirow{6}{*}{\{No. trees, max depth\}} & \{50, \textit{default}\} \\
	    
	    \texttt{ET\_2} & & & \{50, 3\} \\
	    
	    \texttt{ET\_3} & & & \{50, 5\} \\
	    
	    \texttt{ET\_4} & & & \{100, \textit{default}\} \\
	    
	    \texttt{ET\_5} & & & \{100, 3\} \\
	    
	    \texttt{ET\_6} & & & \{100, 5\} \\
	    
	    \midrule   
	    
	    \texttt{KNN\_1} & \multirow{10}{*}{K-nearest neighbors} & \multirow{10}{*}{\{K, weight\}} & \{1, uniform\} \\
	    
	    \texttt{KNN\_2} & & & \{5, uniform\} \\
	    
	    \texttt{KNN\_3} & & & \{10, uniform\} \\
	    
	    \texttt{KNN\_4} & & & \{20, uniform\} \\
	    
	    \texttt{KNN\_5} & & & \{50, uniform\} \\
	    
	    \texttt{KNN\_6} & & & \{1, distance\} \\
	    
	    \texttt{KNN\_7} & & & \{5, distance\} \\
	    
	    \texttt{KNN\_8} & & & \{10, distance\} \\
	    
	    \texttt{KNN\_9} & & & \{20, distance\} \\
	    
	    \texttt{KNN\_10} & & & \{50, distance\} \\
	    
	    \midrule   
	    
	    \texttt{PPR} & Projection pursuit regression &   & \textit{default} \\
        
	    \midrule 
	    
	    \texttt{LASSO\_1} & \multirow{4}{*}{LASSO regression} & \multirow{4}{*}{\{regularization\}} & \{1\}\\
	    
	    \texttt{LASSO\_2} & & & \{0.75\}\\
	    
	    \texttt{LASSO\_3} & & & \{0.5\}\\
	    
	    \texttt{LASSO\_4} & & & \{0.25\}\\
	    
	    \midrule 
	    
	    \texttt{RIDGE\_1} & \multirow{4}{*}{Ridge regression} & \multirow{4}{*}{\{regularization\}} & \{1\}\\
	    
	    \texttt{RIDGE\_2} & & & \{0.75\}\\
	    
	    \texttt{RIDGE\_3} & & & \{0.5\}\\
	    
	    \texttt{RIDGE\_4} & & & \{0.25\}\\

	    \midrule 
        
        \texttt{EN} & Elastic-net regression &  & \textit{default} \\
        
        \midrule   
        
        \texttt{PLS\_1} & \multirow{3}{*}{Partial least squares regression} & \multirow{3}{*}{No. components} & 2 \\
        
        \texttt{PLS\_2} & & & 3 \\
        
        \texttt{PLS\_3} & & & 5 \\
        
        \midrule   
        
        \texttt{PCR\_1} & \multirow{3}{*}{Principal components regression} & \multirow{3}{*}{No. components} & 2 \\
        
        \texttt{PCR\_2} & & & 3 \\
        
        \texttt{PCR\_3} & & & 5 \\
        
		\bottomrule    
	\end{tabular}%
	}
	\label{tab:expertsspecs}
\end{table}

\subsubsection{Forecast Combination Methods}\label{sec:combs}

In terms of forecast combination methods, we focus on the following approaches.

\begin{itemize}
    
    \item \texttt{Simple}: Combination rule which assigns equal weights to all models. In practice, the predictions of the available models are combined using the arithmetic mean;
    
    \item \texttt{Window}: Dynamic weighted average of the predictions of the available models \cite{newbold1974experience}. The weights are computed according to the forecasting performance in the last $\lambda$ observations;
    
    \item \texttt{Blast}: A variant of the \texttt{Window} approach \cite{Blast}. Instead of using past recent performance to weigh the available models, the idea is to select the model with the best performance in the last $\lambda$ observations;
    
    \item \texttt{ADE}: A dynamic combination approach based on a meta-learning strategy called arbitrating \cite{cerqueira2019arbitrage}. The idea is to build a meta model (a Random Forest) for each (base) model in the ensemble. Each meta model is designed to predict the error of the corresponding base model. Then, the models in the ensemble are weighted according to the error forecasts. We refer to the work by Cerqueira et al. \cite{cerqueira2019arbitrage} for a complete read on this method;
    
    \item \texttt{EWA}: A dynamic combination rule based on an exponentially weighted average. This method follows the popular weighted majority algorithm \cite{Cesa-Bianchi:2006:PLG:1137817};
    
    \item \texttt{FS}: The fixed share dynamic combination approach. This method is designed to handle non-stationary time series;
    
    \item \texttt{MLpol}: A dynamic combination method based on a polynomially weighted average;
    
    \item \texttt{Best}: A baseline which selects the individual model in the ensemble with the best performance in the training data to predict all the test instances;
    
    \item \texttt{LossTrain}: Another baseline which weights the available models based on the error on the training set. The weights are static and fixed for all testing observations;
    
\end{itemize}

Most of these combination approaches are dynamic to cope with the non-stationarities present in the time series. The exceptions are \texttt{LossTrain} and  \texttt{Simple}.
We followed the study by Cerqueira et al. \cite{cerqueira2019arbitrage} to set the value of the $\lambda$ parameter to 50 observations.

\subsubsection{Weighting Approaches Over the Horizon}\label{sec:weightfuncs}

Dynamic ensemble methods typically assume immediate feedback. They are designed only for one-step ahead forecasting.
Thus, it is not clear how the ensemble weights should be computed along the forecasting horizon.
We study the following approaches to estimate the weights at each time-step:

\begin{itemize}

\item \texttt{Complete Horizon (CH)}: The weights of individual models are estimated using their average performance over the complete forecasting horizon;

\item \texttt{Individual Horizon (IH)}: The ensemble estimates different weights for each horizon;

\item \texttt{First Horizon Forward (FHF)}: The weights computed for the first horizon are propagated over the rest of the horizon;

\item \texttt{Last Horizon Backward (LHB)}: For completeness, we include the inverse approach to \texttt{FHF}. According to \texttt{LHB}, the weights computed for the last horizon are propagated backward to all horizons before it.

\end{itemize}

We test all these variants with the combination rules presented above. The exception is \texttt{Simple}, whose weights are static and not dependent on forecasting performance. This leads to a total of 33 variants for analysis.

\subsection{Experimental Design}\label{ssec:ed}

We estimate the forecasting performance of models using a Monte Carlo cross-validation procedure \cite{picard1984cross}, which is also referred to as repeated holdout \cite{cerqueira2020evaluating}.
This estimation method is applied with 10 folds. The training and test sizes of each fold are set to 60\% and 10\% of the size of the input time series, respectively. 
Monte Carlo cross-validation provides competitive performance estimates relative to other approaches \cite{cerqueira2020evaluating}.

We preprocess each time series as follows. We take the first differences to remove the trend. Then, we apply time delay embedding to transform the series for auto-regression. We set the parameter $q$ (the number of lags) to 5. This means that the future values of a time series are modeled based on the previous 5 observations.
Finally, we set the maximum forecasting horizon to 18 observations ($H=18$). 

The mean absolute error (MAE) is used as the evaluation metric. This metric has the limitation of being scale-dependent. However, we will focus on ranks and percentage differences to compare results across multiple time series.

\section{Experiments}\label{sec:ee}

This section presents the experiments carried out to analyze the performance of different dynamic ensembles for multi-step ahead forecasting. First, we outline the research questions (Section \ref{sec:exp1}). Then, we answer these questions in turn (Section \ref{sec:exp2}).

\subsection{Research Questions}\label{sec:exp1}

The experiments are designed to address the following research questions:

\begin{itemize}
    \item RQ1: How do the dynamic ensemble methods compare with each other in terms of their rank across all data sets?
    
    \item RQ2: What is the best approach for computing the weights along the forecasting horizon?
    
    \item RQ3: How does each dynamic ensemble method compare with a static ensemble that assigns equal weights to all models?
    
    \item RQ4: Does the forecasting horizon affect the results obtained?
\end{itemize}

\subsection{Results}\label{sec:exp2}

This section answers each research question in turn.

\subsubsection{RQ1}\label{sec:rq1}

\begin{figure}[h]
    \centering
    \includegraphics[width=\textwidth, trim=0cm 0cm 0cm 0cm, clip=TRUE]{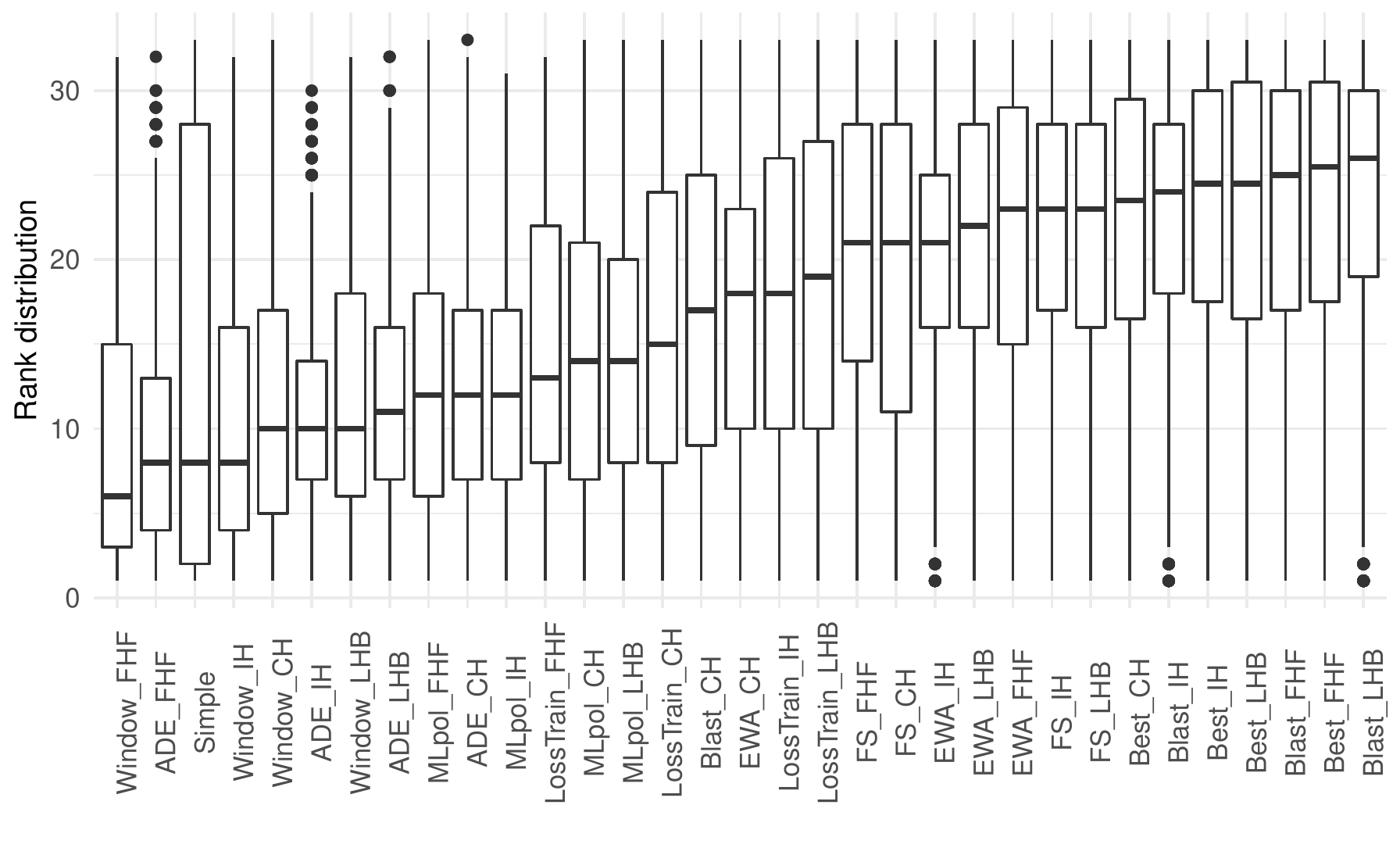}
    \caption{Rank distribution of each method across all time series. Methods are ordered by median rank.}
    \label{fig:0}
\end{figure}

As an exploratory analysis of the results, we start by studying the rank distribution of each approach across the 3568 time series. An approach gets a rank of 1 in a given data set if it shows the lowest error. The average rank quantifies the average position of a method relative to the remaining ones. 
This analysis is presented in Figure \ref{fig:0}. \texttt{Window\_FHF}, \texttt{ADE\_FHF}, and Simple show the top three best median rank. 
The distributions suggest that all methods perform best in some data sets. There is not a single approach that is the best one for all problems. This result aligns with the No Free Lunch theorem for supervised learning put forth by Wolpert \cite{wolpert2002supervised}.

\begin{figure}[h]
    \centering
    \includegraphics[width=\textwidth, trim=0cm 0cm 0cm 0cm, clip=TRUE]{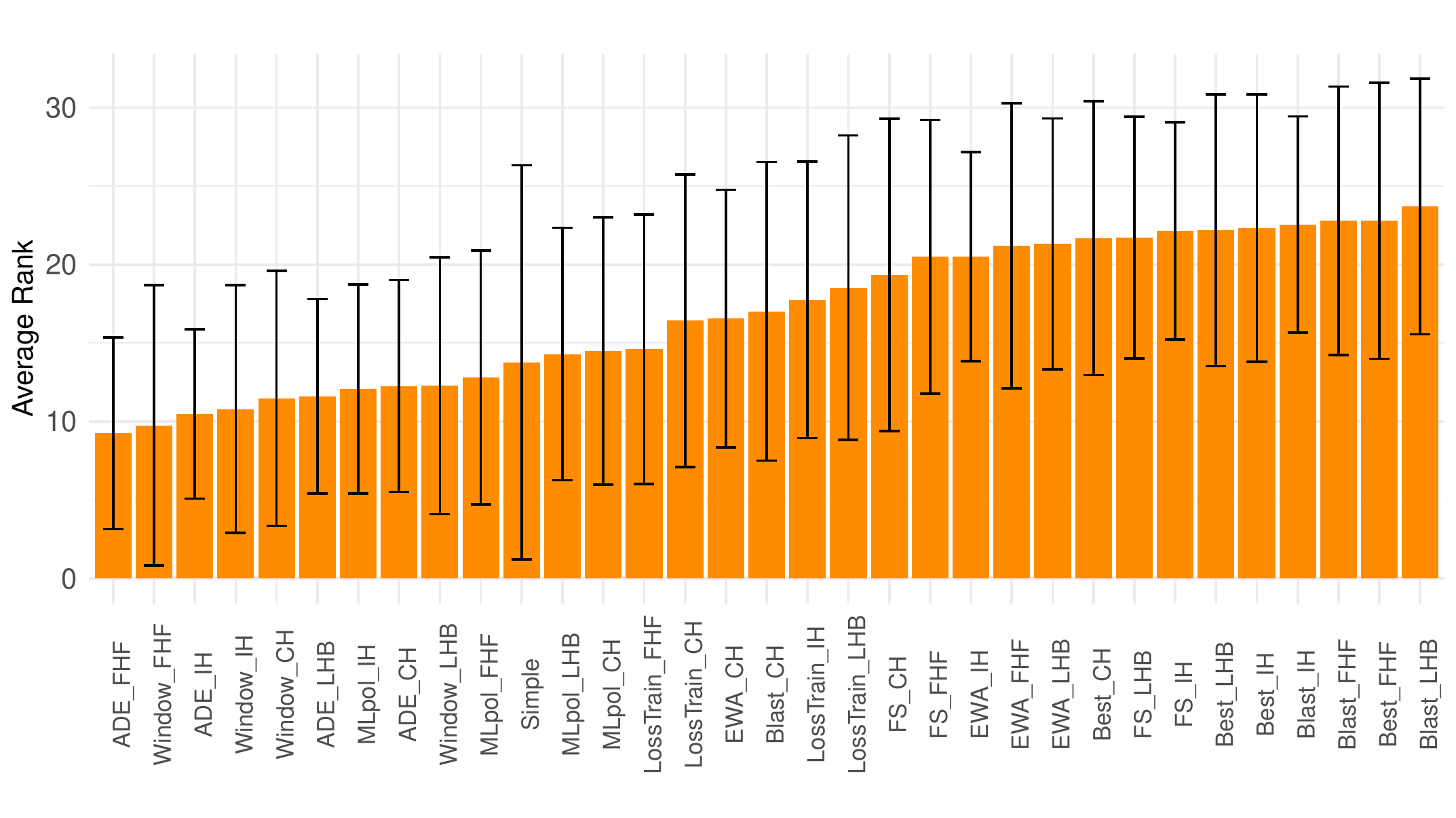}
    \caption{Average rank, and respective standard deviation, of each method across all time series.}
    \label{fig:2}
\end{figure}

Figure \ref{fig:2} shows the average (mean) rank, and respective standard deviation, of each method across the 3568 time series. The best approaches are variants of \texttt{ADE}, \texttt{Window}, and \texttt{MLpol}. On the other hand variants of \texttt{Blast}, \texttt{Best}, \texttt{FS}, and \texttt{EWA} occupy the bottom positions.
All methods show a considerable standard deviation of rank. This corroborates the idea that no method dominates over the rest.

\subsubsection{RQ2}\label{sec:rq2}

\begin{figure}[h]
    \centering
    \includegraphics[width=\textwidth, trim=0cm 0cm 0cm 0cm, clip=TRUE]{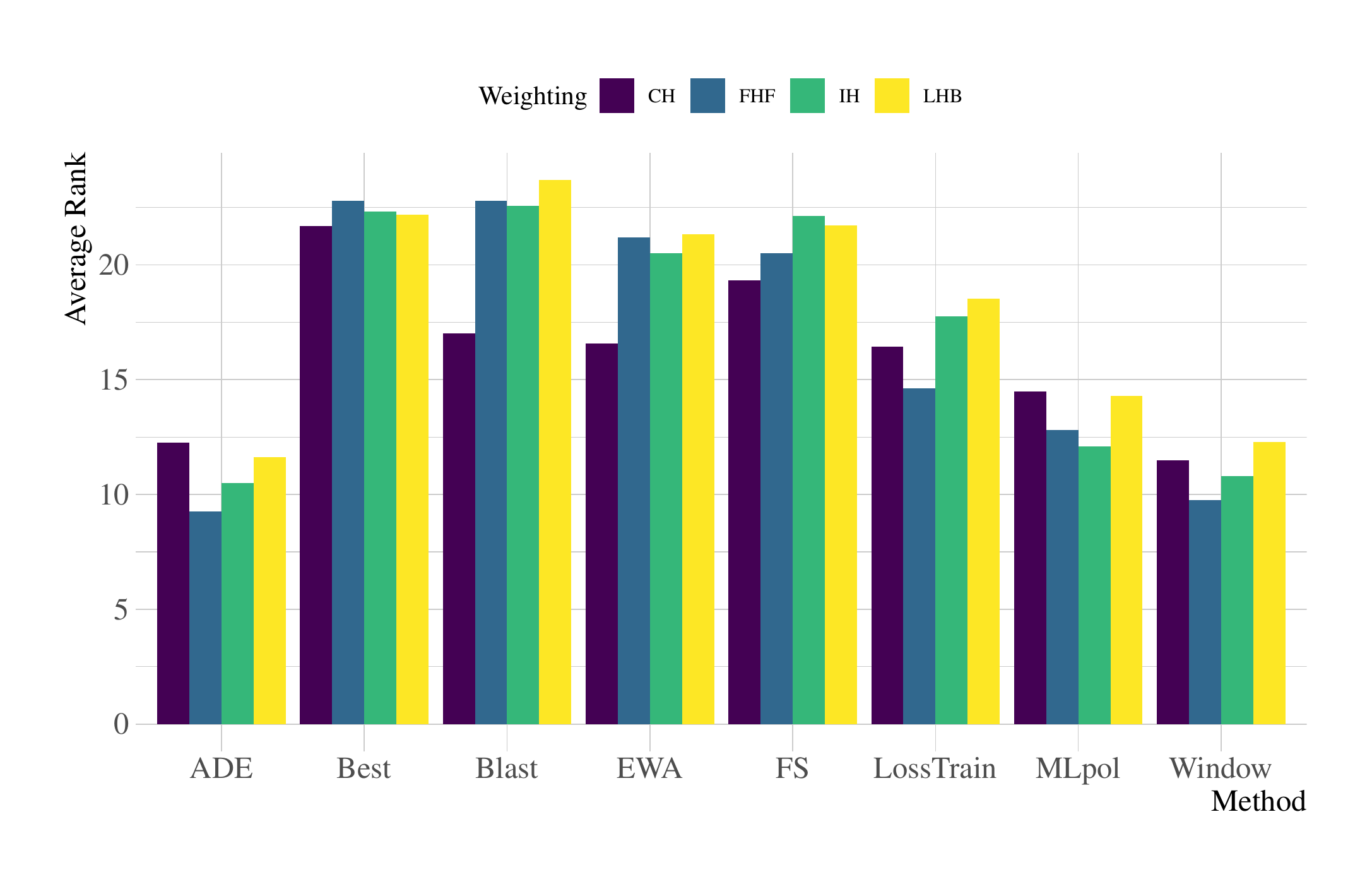}
    \caption{Average rank of each method, but aggregated by weighting strategy.}
    \label{fig:9}
\end{figure}

Figure \ref{fig:9} illustrates the average rank of all methods but aggregated by weighting strategy. This analysis uncovers interesting outcomes. Varying the weighting strategy (\texttt{FHF}, \texttt{LHB}, \texttt{IH}, \texttt{CH}) does not have a significant impact in average rank. On the other hand, different combination approaches show significantly different scores. For example, \texttt{ADE} shows better performance relative to \texttt{Best} irrespective of the weighting strategy. 
The two best forecast combination methods, \texttt{ADE} and \texttt{Window}, show similar behavior in terms of relative scores for the different weighting strategies. Their best score is achieved with \texttt{FHF}, followed by \texttt{IH}. Conversely, the worst combination approaches, namely \texttt{Blast}, \texttt{Best}, \texttt{EWA}, and \texttt{FS}, have their best score when coupled with \texttt{CH}.

\begin{figure}[h]
    \centering
    \includegraphics[width=\textwidth, trim=0cm 0cm 0cm 0cm, clip=TRUE]{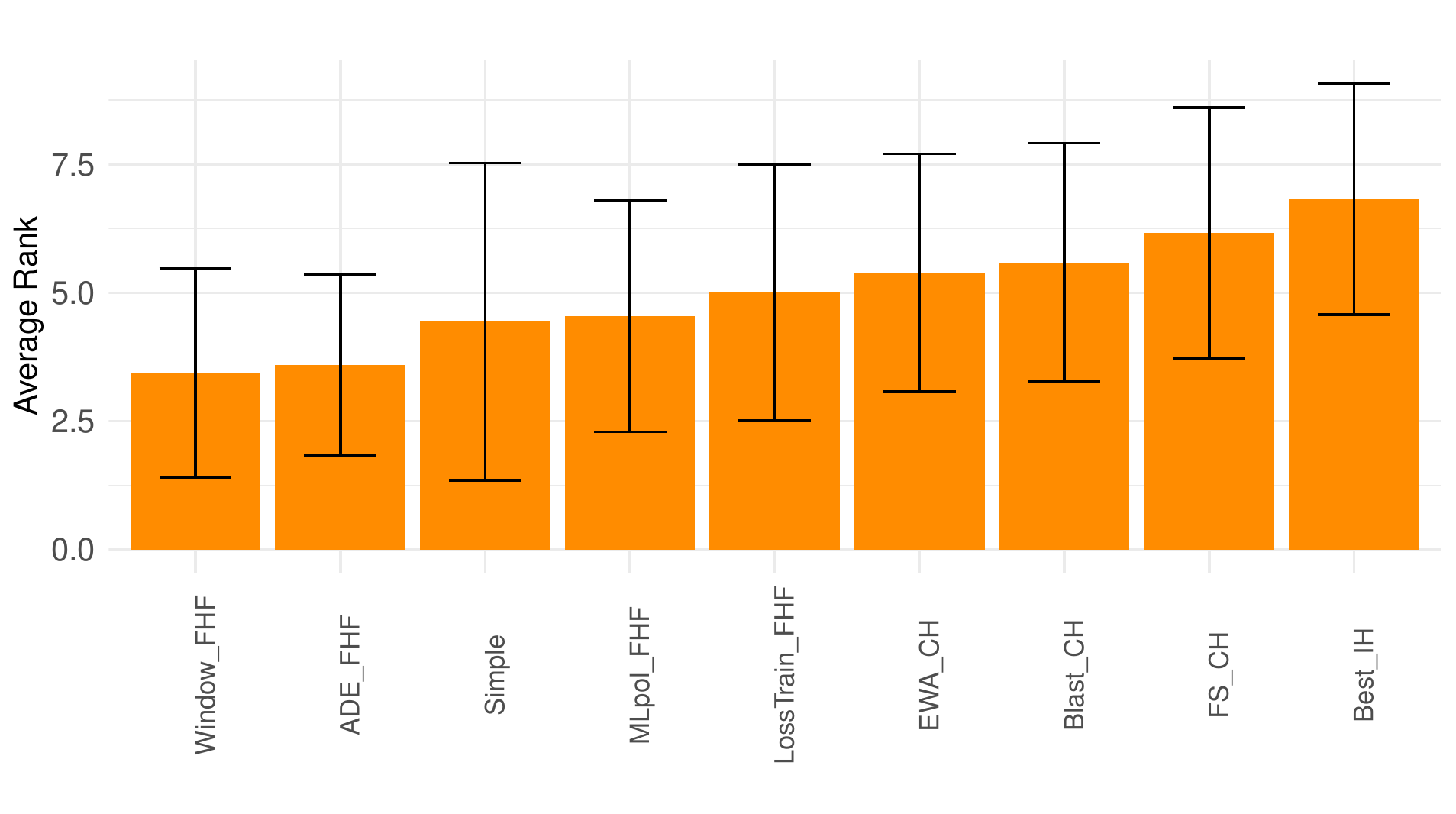}
    \caption{Average rank, and respective standard deviation, of each combination rule applied with the respective best approach for computing the weights across the forecasting horizon.}
    \label{fig:4}
\end{figure}

In Figure \ref{fig:4}, we repeated the average rank analysis but focused on the best version of each combination method. The best weighting scheme for each forecast combination method is detailed in Table \ref{tab:bestw}.

\begin{table}[!bth]
	\centering
	\caption{Best weighting scheme for each combination rule}		
	\begin{tabular}{ll}
	\toprule
	\textbf{Combination Rule} & \textbf{Best Approach} \\
	    \midrule
        
        \texttt{ADE} & \texttt{FHF} \\
        
        \midrule  
        
        \texttt{Best} & \texttt{CH} \\        
        
        \midrule  
        
        \texttt{Blast} & \texttt{CH} \\
        
        \midrule  
        
        \texttt{EWA} & \texttt{CH} \\
        
        \midrule  
        
        \texttt{FS} & \texttt{CH} \\
        
        \midrule
        
        \texttt{LossTrain} & \texttt{FHF} \\
        
        \midrule
        
        \texttt{MLpol} & \texttt{IH} \\
        
        \midrule
        
        \texttt{Window} & \texttt{FHF} \\
        
		\bottomrule    
	\end{tabular}%
	\label{tab:bestw}
\end{table}

The best approach is \texttt{Window\_FHF} followed by \texttt{ADE\_FHF}.
The results showed so far suggest that most methods struggle to outperform the baseline \texttt{Simple}: the approach which combines the individual models with equal weights in all forecasting horizons. Indeed, in Figure \ref{fig:4} \texttt{Simple} shows the third-best average rank.

\subsubsection{RQ3}\label{sec:rq3}

So far, the results were analyzed according to the average rank. However, average rank ignores the magnitude of differences in predictive performance \cite{benavoli2016should}. To overcome this limitation, we also study the percentage difference in performance between each method and a reference method. We set \texttt{Simple} as the reference method, which assigns equal weights to the models in the ensemble.

For each method m, the percentage difference is computed as follows.

\begin{equation*}
    100 \times \frac{\text{MAE}_{\texttt{m}} - \text{MAE}_{\texttt{Simple}} }{\text{MAE}_{\texttt{Simple}}}
\end{equation*}

\noindent where $\text{MAE}_{\texttt{m}}$ and $\text{MAE}_{\texttt{Simple}}$ represent the MAE of method \texttt{m} and \texttt{Simple}, respectively. Negative values denote better performance by method m.

\begin{figure}[h]
    \centering
    \includegraphics[width=\textwidth, trim=0cm 0cm 0cm 0cm, clip=TRUE]{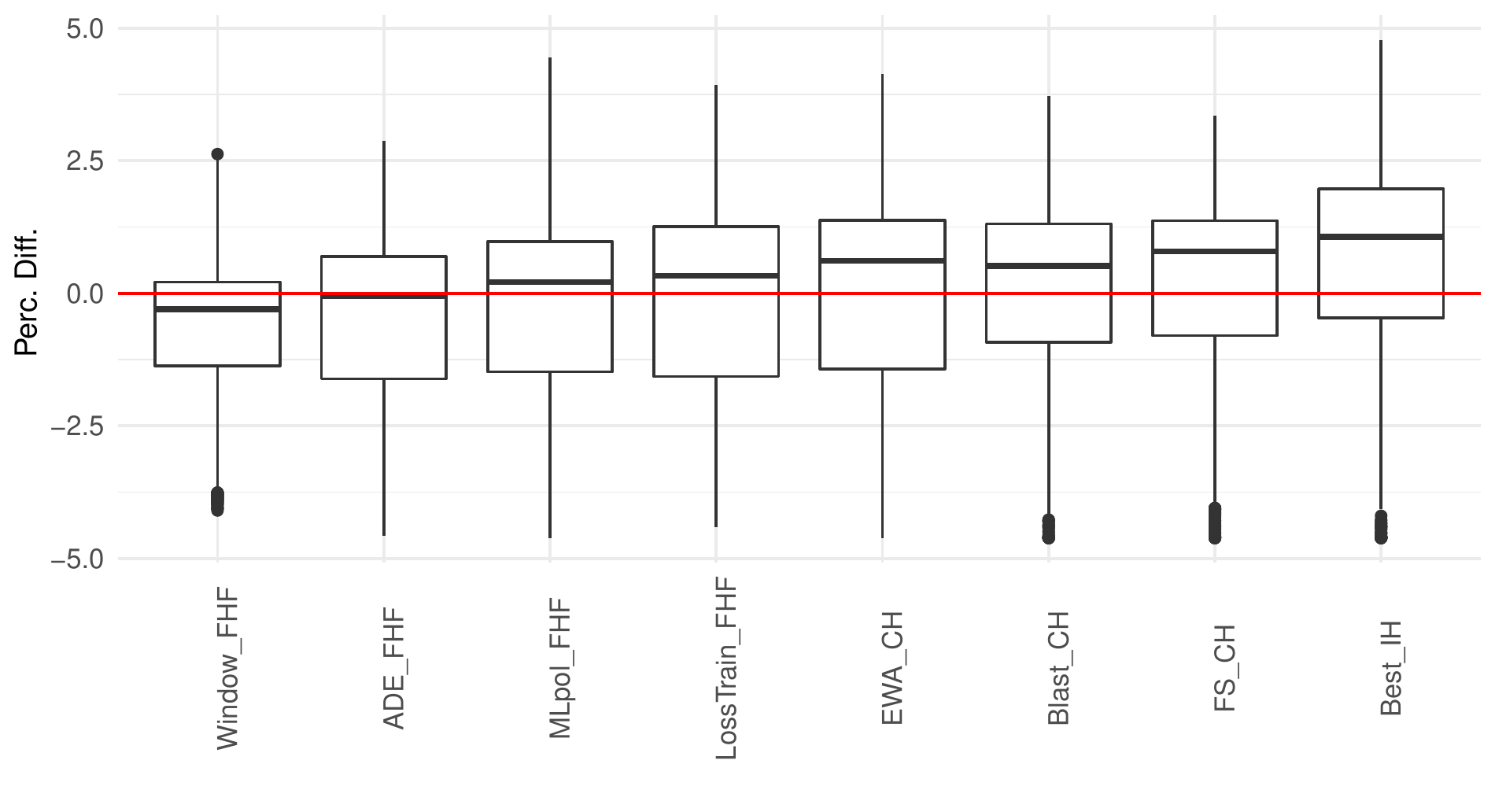}
    \caption{Distribution of percentage difference in MAE between each method and \texttt{Simple} across all time series. Negative values denote better performance of the respective method.}
    \label{fig:5}
\end{figure}

Figure \ref{fig:5} depicts the distribution of the percentage difference in MAE between each method and \texttt{Simple}. The methods are ordered by decreasing median percentage difference in MAE.

The order of the methods is similar to that obtained in the previous analyzes. Only \texttt{ADE\_FHF} and \texttt{Window\_FHF} show a median performance difference below zero. This indicates that \texttt{Simple} outperforms the other dynamic combination methods more times than not.

Figure \ref{fig:5} indicates that the percentage difference is close to zero in many problems.
We analyzed the percentage difference in performance but considered small differences to be negligible.
More specifically, we consider percentage differences below 1\% to be a draw between the pair of methods under analysis.

\begin{figure}[h]
    \centering
    \includegraphics[width=\textwidth, trim=0cm 0cm 0cm 0cm, clip=TRUE]{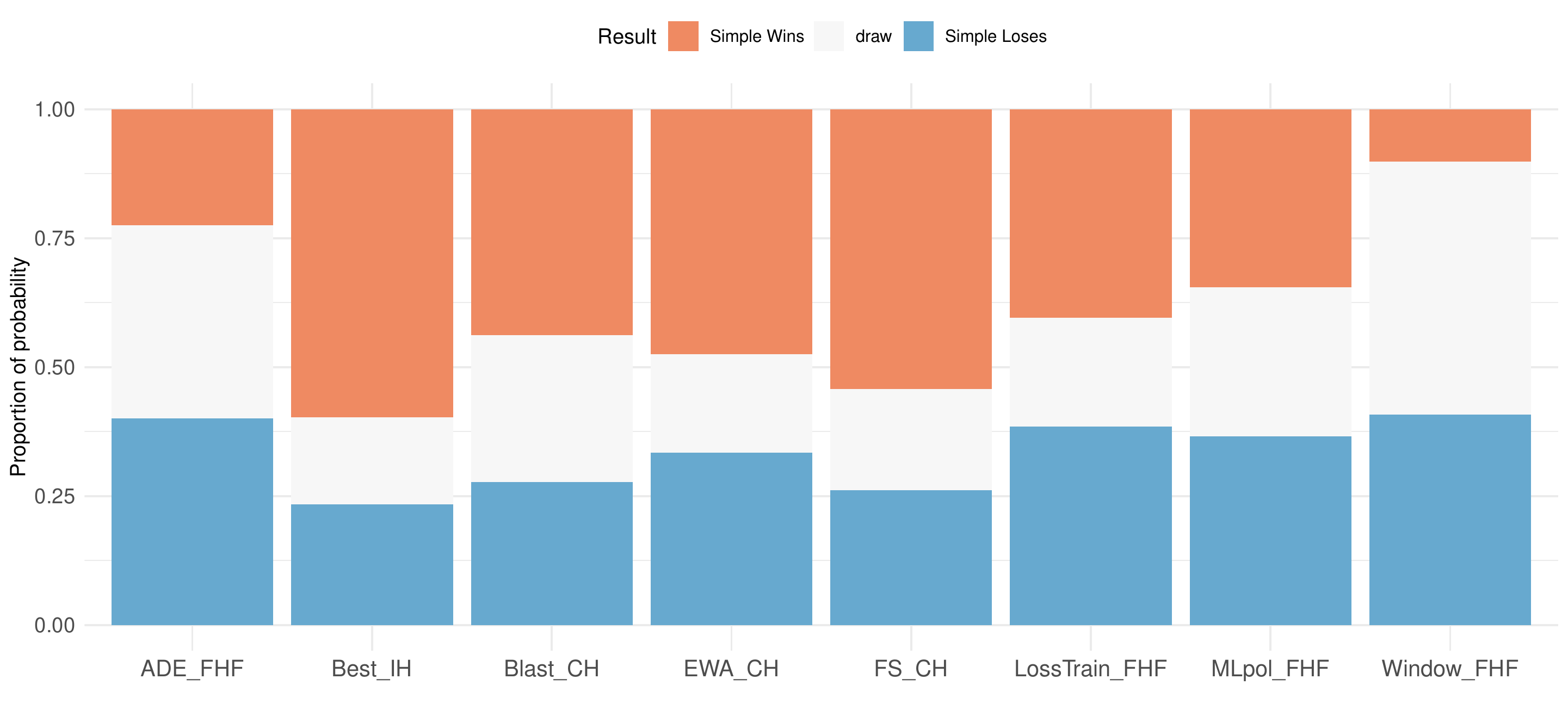}
    \caption{Paired comparisons between each method and \texttt{Simple}. Each barplot shows the probability of each method winning in blue (result below -1\%), drawing in light grey (result within [-1\%, 1\%]), or losing in red (result above 1\%) against \texttt{Simple}.}
    \label{fig:6}
\end{figure}

The outcome of this analysis is illustrated in Figure \ref{fig:6}. 
This figure shows the probability of each method winning in blue (percentage difference below -1\%), drawing in light grey (percentage difference below 1\%), or losing in red (percentage difference above 1\%) against \texttt{Simple}. 
As an example, \texttt{ADE\_FHF} wins in around 40\% of the datasets. A draw occurs in about 35\% of the time series, while \texttt{Simple} wins in around 25\% of the cases.
The scores obtained indicate that only \texttt{ADE\_FHF}, \texttt{Window\_FHF}, \texttt{MLpol\_FHF}, and \texttt{LossTrain\_FHF} are competitive with \texttt{Simple}. When compared with the remaining methods, the probability of \texttt{Simple} winning is considerably higher than the opposite outcome.

\subsubsection{RQ4}

The analysis presented so far quantifies the performance of each method for long-term forecasting. Specifically, predicting 18 step in advance. However, long-term forecasts are typically less accurate than short-term ones.
We analyzed the impact of the forecasting horizon in the results obtained.

\begin{figure}[h]
    \centering
    \includegraphics[width=\textwidth, trim=0cm 0cm 0cm 0cm, clip=TRUE]{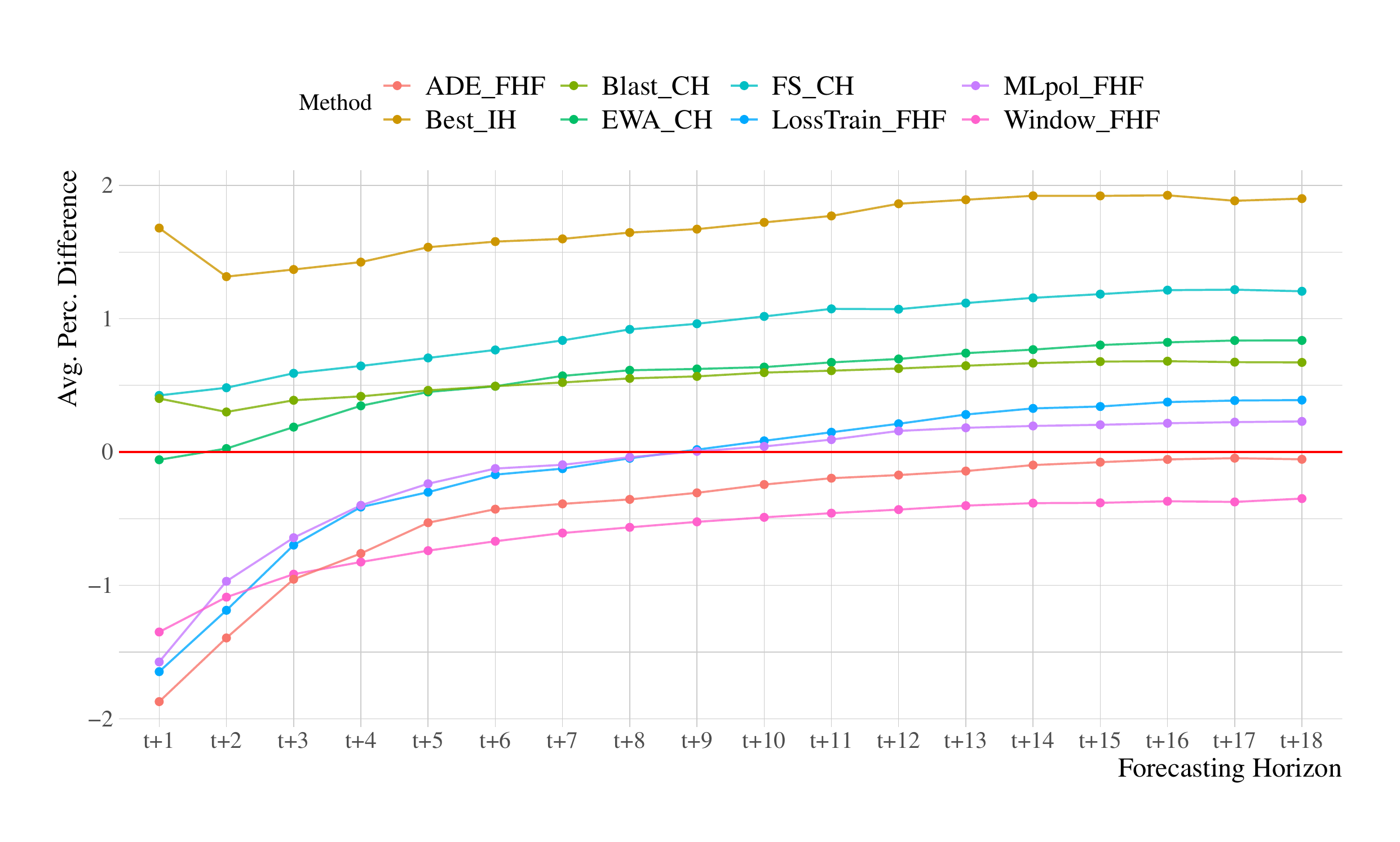}
    \caption{Median percentage difference of each method relative to \texttt{Simple} in each forecasting horizon.}
    \label{fig:7}
\end{figure}

Figure \ref{fig:7} shows the average (median) percentage difference of each method relative to \texttt{Simple} over the forecasting horizon. As before, the average is computed across the 3568 time series. 

The figure shows a clear trend which indicates that the methods decrease their performance relative to Simple as the forecasting horizon increases.
For t+1 (one-step ahead forecasting), 5 out of 8 combination methods outperform Simple. But, for long-term forecasting (t+18), only two methods achieve better performance.
It is also interesting to note that there are slight changes in the relative performance of methods. For example, for t+1 \texttt{Window\_FHF} is only the fourth best approach (\texttt{ADE\_FHF} shows the best average rank). However, by t+18, \texttt{Window\_FHF} shows the best score.

\section{Discussion}\label{sec:disc}

This paper investigates the performance of several dynamic ensembles for multi-step ahead forecasting problems. We focus on ensembles composed of multi-output models. Notwithstanding, the combination rules tested in this work can also be applicable to models following a different strategy regarding multi-step ahead prediction (c.f. Section \ref{sec:rw2.3}).

The main motivation for this work is that the literature concerning dynamic ensembles for forecasting is focused on one-step ahead predictions. Most approaches assume immediate feedback from the environment to compute the error of each method. Previous works concerning dynamic ensembles for multi-step ahead forecasting are scarce. Thus, it is not clear how these methods should weigh the individual models across the forecasting horizon.

We carried out a set of experiments that suggest the following conclusions. All methods show high variability in their rank over the time series used in the experiments. Notwithstanding, variants of \texttt{ADE}, and \texttt{Window} show the best average rank (RQ1).

Four different approaches were tested to compute the ensemble weights across the forecasting horizon. We did not find a significant impact on relative performance between the different approaches. Yet, the methods with the best overall performance maximize their scores when applied with the \texttt{FHF} strategy (RQ2).

The dynamic ensembles were compared with a static ensemble which assigns equal weights to all models. We found that only two methods systematically outperform this approach, namely \texttt{ADE\_FHF}, and \texttt{Window\_FHF} (RQ3).

We also discovered that the forecasting horizon significantly affects the relative performance of dynamic ensembles. All methods  decrease their performance relative to \texttt{Simple} as the forecasting horizon increases (RQ4).

Future research opportunities include analyzing how dynamic ensembles can cope with an increasing forecasting horizon, and remain competitive with static combination rules.

\section{Conclusions}\label{sec:fr}

This work addresses multi-step ahead forecasting problems using ensembles composed of multi-output models. A substantial amount of work have studied how dynamic ensembles perform in forecasting tasks. Yet, the literature is focused on one-step ahead predictions, ignoring multi-step ahead problems.

The goal of this work was to bridge this gap in the literature. The main contribution is an extensive empirical study that analyzes how different dynamic ensembles perform for multi-step ahead forecasting. We also study what is the best approach for weighting the models composing the ensemble along the forecasting horizon.

The dynamic methods \texttt{ADE} and \texttt{Window} present the best average rank. On the other hand, other approaches struggle to outperform a static ensemble which assigns equal weights to all models (\texttt{Simple}).
We found that different approaches for computing the ensemble weights along the horizon did not affect performance significantly. Yet, propagating the weights computed for one-step ahead predictions along the horizon maximized the performance of the best dynamic ensembles.
Another important conclusion from this work is that the forecasting horizon has a significant impact on the results obtained. Most dynamic ensembles struggle to outperform \texttt{Simple} as the forecasting horizon increases.

\bibliographystyle{spmpsci.bst} 

\end{document}